\documentclass[letterpaper, 10 pt, conference]{ieeeconf}  

\IEEEoverridecommandlockouts                              
\usepackage{graphics} 
\usepackage{graphicx} 
\usepackage{subcaption} 
\usepackage{amssymb}  
\usepackage{amsmath} 
\usepackage{todonotes}
\usepackage[ruled,vlined]{algorithm2e}
\usepackage{booktabs}
\usepackage{float}
\usepackage[inkscapelatex=false]{svg}

\title{ \LARGE \bf Terrain-Aware Autonomous Planetary Exploration for Exteroceptive-Proprioceptive Mapping with Quadruped Scouts }

\author{
Alberto Sanchez-Delgado$^{*}$, João Carlos Virgolino Soares, Victor Barasuol, Claudio Semini%
\thanks{This work was supported by the Italian Space Agency (ASI).}
\thanks{All authors are with the Dynamic Legged Systems lab (DLS), Istituto Italiano di Tecnologia (IIT), Genoa, Italy. \texttt{\{first\_name.first\_surname\}@iit.it}}%
\thanks{*Corresponding author: \texttt{alberto.sanchez@iit.it}.}
}

\begin{document} 

\maketitle

\begin{abstract}
Autonomous planetary exploration requires robots to navigate unknown, uneven terrain while assessing risk, traversability, and energetic cost. Quadruped scouts are well suited for this task because they can traverse irregular surfaces and gather mobility-relevant information during locomotion. This paper presents a terrain-aware exploration framework that combines exteroceptive and proprioceptive mapping for a quadruped robot in lunar-like environments. An onboard RGB-D camera builds robot-centered elevation maps, estimates geometric traversability, and derives navigation costs for autonomous planning. In parallel, proprioceptive measurements provide interaction-aware terrain cues that complement geometry-based assessment. Local maps are incrementally registered into a global multi-layer representation, which is used by an exploration module to select targets in unexplored regions of interest. The targets are reached by an autonomous navigation system that guides collision-aware motion using the available map and cost layers. Simulation results on NVIDIA Isaac Sim show autonomous exploration, map expansion, and spatial association between terrain geometry and robot-terrain interaction. Subsequent navigation using this information exhibits lower average Cost of Transport (CoT) than initial exploration.
\end{abstract}



\section{Introduction}

The renewed interest in planetary exploration has placed the Moon at the center of future robotic and human missions. Polar and near-polar regions are particularly relevant due to their scientific value, potential access to volatiles, and favorable exploration conditions in selected areas~\cite{li2018waterIce,speyerer2013illuminated,glaser2017illumination}. They are also candidate sites for sustained surface operations and infrastructure~\cite{deRosa2012lunarLander}. However, these regions remain difficult to traverse and assess locally because of rocks, craters, slopes, loose regolith, illumination changes, and extended shadows. These conditions motivate robotic scouts capable of incrementally mapping terrain properties beyond what can be inferred from orbital data or prior Digital Elevation Models (DEMs) alone~\cite{chien2024spaceRobotics,santra2024riskAware,hong2022illuminationVariant}.

Historically, planetary exploration has relied mostly on wheeled rovers, which can face severe mobility limitations such as high slippage or entrapment on loose granular soils~\cite{thoesenPlanetary2021}. Current planetary robotic systems rely on prior maps for global route planning and onboard exteroceptive sensing for local traversability assessment and navigation. Although cameras and depth sensors provide essential geometric information, they do not directly measure the robot's physical response while walking. Proprioceptive methods complement this information by using joint torques and velocities, inertial measurements, and foot-contact signals to characterize how terrain affects the robot's locomotion. This is critical in unstructured planetary terrain, where geometrically similar areas may induce different energetic costs, stability margins, or slippage behavior. Legged robots are promising scouting platforms because they can adapt contacts to irregular surfaces, negotiate steep or discontinuous terrain, and collect mobility-relevant information while moving through the environment~\cite{sanchez2025astra,Valsecchi2023,varadharajan2025multiRobot}.

Recent work has shown the potential of quadrupeds for planetary analog exploration and long-range autonomy in challenging environments~\cite{arm2023,bouman2020autonomousSpot,kulkarni2021autonomousTeamedExploration}. In this direction, we previously introduced a proprioceptive terrain mapping framework to associate locomotion-derived metrics with traversed terrain, including Cost of Transport (CoT), slippage, and Gravito-Inertial Acceleration (GIA)-based stability~\cite{sanchez2025isparo}. However, that research focused on mapping executed trajectories and did not close the loop between terrain assessment, autonomous target generation, and navigation.

This paper extends our previous work~\cite{sanchez2025isparo} by integrating exteroceptive terrain assessment, proprioceptive terrain mapping, and autonomous exploration in a unified framework for a quadruped scout. An RGB-D camera is used to generate robot-centered elevation maps, from which geometric traversability is estimated and converted into navigation costs for autonomous planning. In parallel, proprioceptive measurements estimate CoT, GIA-based stability, and slippage, which are spatially accumulated as interaction-aware layers in a global multi-layer gridmap. Without assuming a prior global terrain map, the robot incrementally builds this representation while exploring and selecting navigation targets toward unknown regions. Our framework supports autonomous exploration in lunar-like environments while associating visible terrain geometry with physical-interaction information across different data types in a multi-layer map.

\subsection{Related Work}

Legged robots have gained attention as high-mobility platforms for exploration in planetary analog sites and unstructured terrain. Compared to wheeled systems, quadrupeds offer high agility in hazardous zones by dynamically selecting discrete contact points for walking. This morphological advantage allows them to traverse steep slopes and rocky surfaces that are otherwise inaccessible to traditional wheeled rovers, making them suitable scouts for evaluating risky terrain before less agile or payload-sensitive rovers are deployed. Arm et al.~\cite{arm2023} demonstrated quadruped robots as part of a team for scientific exploration, while Valsecchi et al.~\cite{Valsecchi2023} studied quadrupedal locomotion on steep planetary terrain. Beyond planetary scenarios, autonomous legged systems have enabled long-range exploration in extreme and communication-constrained environments, further supporting the potential of quadrupeds as scouts~\cite{bouman2020autonomousSpot,kulkarni2021autonomousTeamedExploration}. However, these works mainly emphasize mobility, exploration autonomy, or mission-level coordination, rather than terrain maps that jointly encode geometric traversability and physical robot-terrain interaction.

Exteroceptive terrain mapping is central for navigation in rough environments. Elevation mapping methods provide local geometric representations from depth or LiDAR data for foothold evaluation and obstacle avoidance~\cite{miki2022elevation}. Multimodal elevation maps extend this representation with additional perceptual layers, supporting richer planning and learning-based navigation~\cite{erni2023mem}. In planetary robotics, risk-aware mapping and planning have considered elevation and environmental uncertainty for lunar surface navigation~\cite{santra2024riskAware,hong2022illuminationVariant}. While these methods provide valuable terrain structure, their costs are generally derived from external perception or prior models, and do not directly encode how difficult, unstable, or energetically demanding the terrain is during traversal.

Proprioceptive methods complement exteroception by estimating terrain properties from locomotion signals~\cite{wilson2026legged}. Slip detection, terrain classification, and proprioceptive traversability estimation have been studied for quadrupeds using joint, contact, inertial, or motion-based measurements~\cite{nisticoSlip2022,elnoor2024pronav,cai2025pietra}. GIA-based stability metrics have also been used to assess climbing capability and dynamic equilibrium in challenging conditions~\cite{unoSimulationBased2022}. However, these approaches do not integrate proprioceptive terrain maps with autonomous exploration and navigation.

Autonomous exploration methods generate targets toward unknown or information-rich regions while incrementally extending an environment map. In space robotics, multi-robot planners are designed to coordinate fleet operations, such as groups of traditional wheeled rovers, under intermittent connectivity and operational constraints~\cite{varadharajan2025multiRobot}. Meanwhile, although teams of legged rovers and  lunar drones have not yet been deployed in actual space missions, recent research in terrestrial analogues, including underground mines, demonstrates that their integrated onboard mapping and frontier selection can enhance autonomous exploration~\cite{kulkarni2021autonomousTeamedExploration}. 

More recently, Jiang et al. estimated safe and frontier regions online from leg–terrain interactions and paired them with a reactive controller for safe exploration of granular terrain, and used multi-objective frontier selection for safe goal-directed navigation~\cite{jiang2026proprioceptive}. These frameworks, however, drive exploration from proprioceptively estimated traversability alone, without registering the geometric structure of the terrain using exteroceptive sensors. Conversely, exteroceptive multi-layer maps capture geometric or semantic risk sources but not the physical interaction experienced during traversal~\cite{bouman2020autonomousSpot}.

In contrast, this work fuses both: a quadruped scout incrementally builds a global multi-layer map that co-registers exteroceptive geometric traversability with proprioceptive interaction layers, encoding how difficult, unstable, or energetically demanding the terrain is to traverse. Since traversability depends on rover morphology, including size, weight, and locomotion type, these interaction-aware maps could support trafficability assessment for heterogeneous fleets, subject to platform-specific adaptation and validation.

\subsection{Main Contributions}
This paper presents an exploration framework for a quadruped scout operating in lunar-like environments. Building on previous work on proprioceptive terrain mapping~\cite{sanchez2025isparo}, the proposed system integrates exteroceptive terrain assessment, proprioceptive terrain layers, and autonomous target generation within a unified exploration pipeline. The main contributions are:

\begin{itemize}
\item \textbf{Exploration-driven exteroceptive–proprioceptive terrain framework:} We propose a framework in which a quadruped scout, without a prior global map, autonomously explores unknown terrain and incrementally builds a global multi-layer representation that co-registers elevation and geometric traversability with proprioceptive interaction metrics (CoT, GIA-based stability, slippage), spatially associating visible geometry with the physical interaction experienced during locomotion.

\item \textbf{Integrated evaluation in lunar-like exploration scenarios:} We evaluate the complete pipeline in photorealistic simulation and compare initial exploration with subsequent navigation over the same target sequence. The latter reuses the accumulated terrain information and exhibits lower average CoT.
\end{itemize}



\section{Exploration and Mapping Framework}

\begin{figure*}[htbp]
    \centering
    \includegraphics[width=0.90\linewidth]{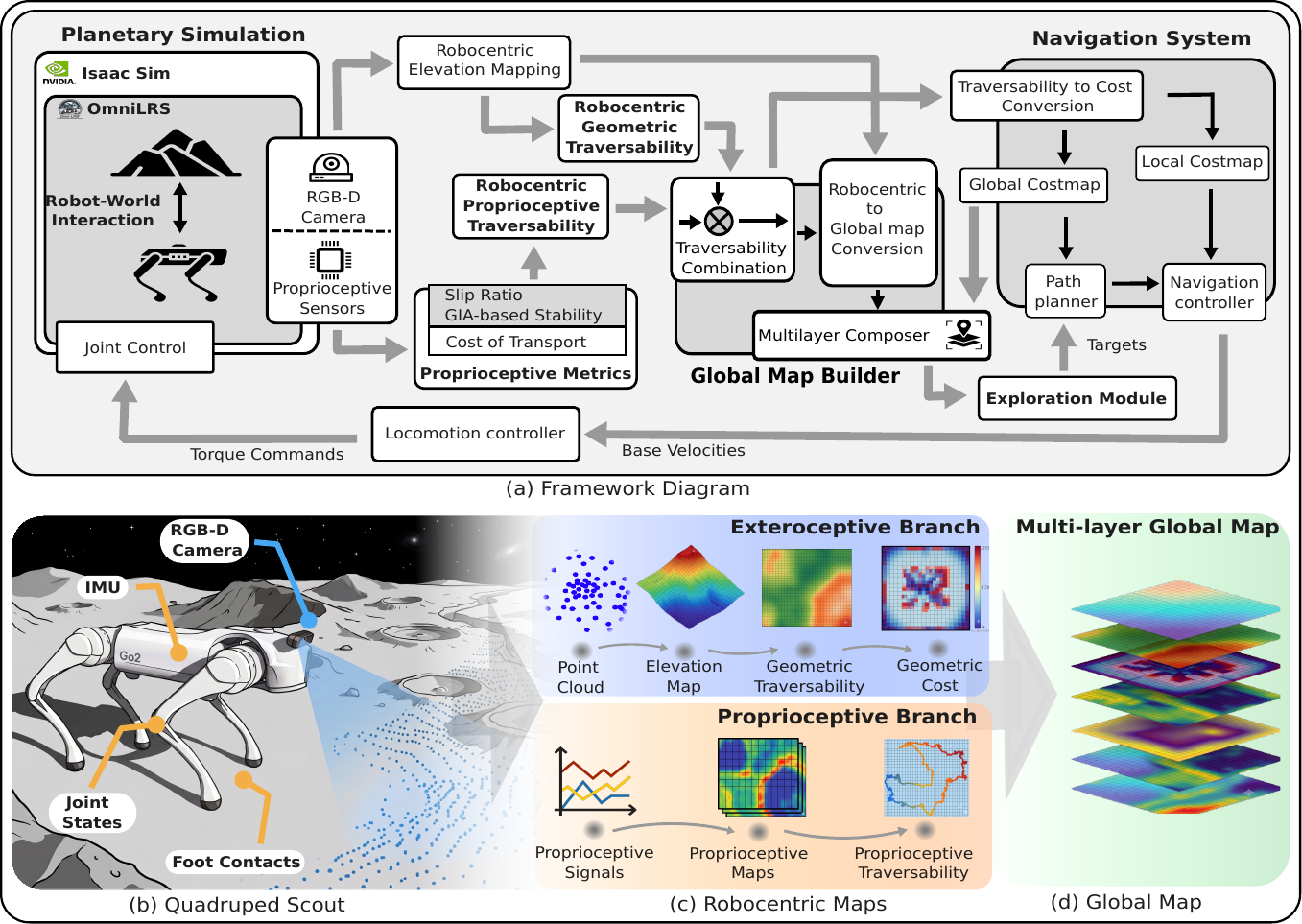}
    \caption{Overview of the proposed exploration and mapping framework. 
        (a) System architecture. 
        (b) RGB-D and proprioceptive sensing on lunar-like terrain. 
        (c) Conversion of exteroceptive and proprioceptive measurements into robot-centered terrain maps. 
        (d) Global multi-layer gridmap. (AI usage: Some parts of subfigures (b) and (d) were generated by AI.)}
    \label{fig:framework}
\end{figure*}

This section describes the proposed framework for a quadruped scout operating in initially unknown planetary-like terrain. The objective is to build, online and without a prior global map, a persistent representation that supports autonomous navigation while storing both terrain geometry and robot-terrain interaction. As shown in Fig.~\ref{fig:framework}, RGB-D point clouds provide the information for generating elevation, geometric traversability, and navigation-cost layers, while proprioceptive signals are used to estimate slippage, CoT, and GIA-based stability. These exteroceptive and proprioceptive observations are spatially represented into a global multi-layer representation, built upon the gridmap framework~\cite{fankhauser2016gridmap}, which supports exploration-target generation and terrain-aware navigation. While sub-sections \ref{sec:ext_ele_map},  \ref{sec:prop_terr_map}, \ref{mult-glob-map}, \ref{aut-ext-nav} review standard formulations from the literature, sub-section \ref{ext-prop-trav} describes our specific and novel approach to integrating these distinct information layers.

\subsection{Exteroceptive Elevation Mapping and Traversability Estimation} \label{sec:ext_ele_map}

The exteroceptive branch starts by building an elevation map as proposed in~\cite{miki2022elevation,erni2023mem}. RGB-D measurements are converted into a local point cloud, transformed using the estimated robot pose, and stored in a 2.5D elevation grid (see Fig.~\ref{fig:framework}b-c), with a cell size of $0.1~\mathrm{m}\times0.1~\mathrm{m}$, used consistently across all mapping stages.. For each incoming point associated with a grid cell $j$, the elevation update is computed according to:
\begin{equation}
h_t^j =
\frac{\sigma_p^2 h_{t-1}^j + \sigma_{m,t-1}^2 z_p}
{\sigma_{m,t-1}^2+\sigma_p^2},
\label{eq:elevation_update}
\end{equation}
where $h_t^j$ and $h_{t-1}^j$ are the updated and previous terrain height values associated with the map cell $j$ (in meters), $z_p$ is the measured point height (in meters), and $\sigma_{m,t-1}^2$ and $\sigma_p^2$ denote the map and measurement variances, respectively. Geometric traversability is estimated through grid-map filters applied to the local elevation map. Missing height values are first inpainted via spatial interpolation to obtain $h_{\mathrm{inp}}$.  A smoothed elevation layer, named $h_{\mathrm{sm}}$, is then computed by applying a mean filter to $h_{\mathrm{inp}}$, and surface normals are estimated from the inpainted elevation. The local slope is computed from the vertical component of the terrain normal according to:
\begin{equation}
s(i,j)=\arccos\left(n_z(i,j)\right),
\label{eq:slope}
\end{equation}
where $n_z(i,j)$ is the vertical component of the terrain normal at cell $(i,j)$. Note that since the surface normal is a unit vector ($\Vert\hat{n}\Vert=1$), its vertical component uniquely determines the magnitude of the terrain's inclination angle (slope) relative to the horizontal plane, independently of the slope's azimuth or direction. In parallel, roughness is approximated from the absolute difference between the inpainted and smoothed elevation layers using:
\begin{equation}
r(i,j)=\left|h_{\mathrm{inp}}(i,j)-h_{\mathrm{sm}}(i,j)\right|,
\label{eq:roughness}
\end{equation}
where $h_{\mathrm{inp}}$ and $h_{\mathrm{sm}}$ denote the inpainted and smoothed elevation layers, respectively. The slope and roughness descriptors in~\eqref{eq:slope} and~\eqref{eq:roughness} are combined into a normalized geometric traversability value using:
\begin{equation}
T_g(i,j)=\exp\left[-\left(
\left(\frac{s(i,j)}{s_0}\right)^2+
\left(\frac{r(i,j)}{r_0}\right)^2
\right)\right],
\label{eq:geometric_traversability}
\end{equation}
where $s_0$ and $r_0$ regulate the sensitivity to slope and roughness, respectively. In this work, we set $s_0 = r_0 = 0.5$ to balance the contribution of both normalized parameters. The resulting value $T_g(i,j)\in[0,1]$ represents the geometry-based traversability of each local terrain cell, with higher values indicating smoother and less inclined terrain.

\subsection{Proprioceptive Terrain Mapping} \label{sec:prop_terr_map}

In parallel, proprioceptive sensing provides interaction-aware terrain information. Following~\cite{sanchez2025isparo}, slippage, energetic cost, and GIA-based stability are computed from joint states, contact information and inertial sensing. For each foot \(f\) in stance, slippage is detected from velocity and position discrepancies. Let \(\Delta V_f\) and \(\Delta P_f\) denote the normalized velocity deviation and the position discrepancy between the desired and measured foot motion, respectively. A slip event is then identified according to:
\begin{equation}
    \mathrm{slip}_f \Leftrightarrow
    \Delta V_f>\epsilon_v \;\land\; \Delta P_f>\epsilon_p ,
    \label{eq:slip_event}
\end{equation}
where \(\epsilon_v\) and \(\epsilon_p\) are the corresponding velocity and position thresholds. This criterion associates slippage with simultaneous deviations in foot motion and contact position. The energetic demand of locomotion is represented by the Cost of Transport (CoT), computed from joint torques and velocities according to:

\begin{equation}
    \mathrm{CoT} =
    \frac{1}{mgd}
    \int_{t_0}^{t_f}
    \sum_{k=1}^{N}
    \left|\tau_k(t)\dot{q}_k(t)\right|\,dt ,
    \label{eq:cot}
\end{equation}

where \(m\) is the robot mass, \(g\) is the gravitational acceleration, \(d\) is the robot traveled distance, and \(\tau_k,\dot{q}_k\) are the torque and angular velocity of joint actuator \(k\), respectively. Negative power and energy restoring were not considered in this work.  Stability is evaluated using the Gravito-Inertial Acceleration (GIA)~\cite{unoSimulationBased2022}. In this work, the GIA is approximated from the estimated Center of Mass (CoM) acceleration according to:

\begin{equation}
    a_{GIA}
    =
    g-\ddot{p}_{CoM}(t),
    \label{eq:gia}
\end{equation}

where $\ddot{p}_{\mathrm{CoM}}$ is obtained from the estimated CoM state (computed in practice using Pinocchio~\cite{pinocchioweb,carpentier2019pinocchio}). The resulting GIA vector is evaluated with respect to the current support contacts to obtain stability margins. Finally, a proprioceptive traversability score is obtained by linearly combining normalized slippage and stability, as shown in:

\begin{equation}
    T_p(i,j)=w_s(1-\bar{\rho}_{s}(i,j))+w_\gamma\bar{\gamma}(i,j),
    \label{eq:prop_trav}
\end{equation}

where \(\bar{\rho}_{s}\) is the normalized slip ratio derived from detected slip events, \(\bar{\gamma}\) is the normalized stability margin computed from $a_{\mathrm{GIA}}$ and the support contacts, and \(w_s+w_\gamma=1\). CoT is evaluated separately and does not enter $T_p$. Lower values indicate terrain associated with higher slip or reduced stability. CoT is stored as a separate energy-assessment layer and does not contribute to the combined traversability or navigation cost.

\subsection{Exteroceptive-Proprioceptive Traversability Combination} \label{ext-prop-trav}

The geometric traversability $T_g$ is available before contact, while the proprioceptive traversability $T_p$ is only defined on traversed cells. We therefore use an experience-aware combination, where missing proprioceptive values leave the geometric estimate unchanged. Where $T_p$ is valid, the proprioceptive reward $r_p$ and penalty $q_p$ quantify deviations above and below the neutral value $T_{p,0}$ and are computed using \eqref{eq:prop_reward} and \eqref{eq:prop_penalty}, respectively:
\begin{equation}
\label{eq:prop_reward}
r_p=\max(T_p-T_{p,0},0),
\end{equation}
\begin{equation}
\label{eq:prop_penalty}
q_p=\max(T_{p,0}-T_p,0).
\end{equation}
The combined traversability is then obtained from:
\begin{equation}
\label{eq:combined_trav}
T_c=
\begin{cases}
clip(T_g+\beta_+ r_p-\beta_- q_p,0,1),
& T_p \neq{ NaN},\\
T_g, & \text{otherwise}.
\end{cases}
\end{equation}
Here, $T_{p,0}$ defines a neutral proprioceptive traversability level. Thus, $r_p$ and $q_p$ represent favorable and unfavorable deviations from this reference, respectively. The gains $\beta_+$ and $\beta_-$ weight the influence of proprioceptive feedback on the geometric estimate. The continuous combined traversability value $T_c(i,j)\in[0,1]$ is converted by the traversability-to-costmap block in Fig.~\ref{fig:framework} into an 8-bit navigation cost using:
\begin{equation}
\label{eq}
C_{\mathrm{nav}}(i,j)=\left\lfloor (2^8-3)\left(1-T_c(i,j)\right)\right\rfloor .
\end{equation}
Valid terrain cells are scaled to an 8-bit range ($0$--$252$) to be represented in a 2D costmap. 

\subsection{Multi-layer Global Map Construction} \label{mult-glob-map}

The local terrain representations are accumulated into a persistent global gridmap. At each update, robot-centered observations are transformed to the global frame using the estimated robot pose, as in:
\begin{equation}
    \mathbf{p}^{G} =
    \mathbf{T}^{G}*{R}(t)
    \begin{bmatrix}
    \mathbf{p}^{R} \
    1
    \end{bmatrix},
    \label{eq:local_global_projection}
    \end{equation}
    where $\mathbf{T}^{G}*{R}(t)$ maps the robot frame to the global frame. The grid cell $(i,j)$ is obtained from $\mathbf{p}^{G}$ as in:
    \begin{equation}
    i = \left\lfloor \frac{x^{G}-x_0}{r} \right\rfloor,
    \qquad
    j = \left\lfloor \frac{y^{G}-y_0}{r} \right\rfloor ,
    \label{eq:global_grid_index}
\end{equation}
where $r$ is the map resolution and $(x_0,y_0)$ is the map origin.

The map stores exteroceptive layers, such as elevation and geometric traversability, together with proprioceptive terrain layers. Exteroceptive updates may cover visible but untraversed terrain, whereas proprioceptive updates are only assigned where robot-terrain interaction occurs. For each continuous-valued layer $k$, repeated observations assigned to the same cell are integrated incrementally according to:
\begin{equation}
\phi_{k,t}^{(i,j)}
=
\frac{n_k^{(i,j)}\phi_{k,t-1}^{(i,j)}+\hat{\phi}*{k,t}}
{n_k^{(i,j)}+1},
\label{eq:cell_update}
\end{equation}
where $\phi*{k,t}^{(i,j)}$ is the updated value of layer $k$, $\hat{\phi}_{k,t}$ is the current observation, and $n_k^{(i,j)}$ is the number of previous observations assigned to that cell. The resulting global representation provides the terrain memory used by the exploration and navigation modules.


\subsection{Autonomous Exploration and Navigation} \label{aut-ext-nav}

As shown in Fig.~\ref{fig:framework}, the exploration and navigation modules close the loop between the global map, the selected exploration target, and the locomotion controller. Exploration follows a frontier-based strategy adapted to the assigned Region of Interest (RoI) and to the traversability-aware cost representation, following the principle of selecting targets at the boundary between explored and unknown space~\cite{han2022autoexplorer}. A RoI is a previously unknown square terrain area selected by the mission as a target region that must be explored by the scout. Frontier cells are extracted as the boundary between explored free space and unknown terrain inside the RoI. Cells in relatively higher-cost sub-regions of the RoI or outside it are discarded using \(C_{\mathrm{nav}}\), and the remaining frontier cells define the candidate target set \(\mathcal{F}_{\mathrm{RoI}}\).

Given the current robot state \(q_t\), each candidate target \(\tau \in \mathcal{F}_{\mathrm{RoI}}\) is evaluated with the traversability-aware path cost in:
\begin{equation}
    J_{\mathrm{nav}}(q_t,\tau) =
    \min
    \sum_{(i,j)\in \pi}
    \left(
    \lambda_d \Delta s_{ij}
    +
    \lambda_c C_{\mathrm{nav}}(i,j)
    \right),
    \label{eq:path_cost}
\end{equation}
where \(\Delta s_{ij}\) is the incremental path length, and \(\lambda_d\), \(\lambda_c\) weight distance and terrain difficulty. Thus, \eqref{eq:path_cost} assigns lower cost to shorter paths that cross more traversable terrain. The next exploration target $\tau^\star$ is obtained from:

\begin{equation}
    \tau^\star = \arg\min_{\tau \in \mathcal{F}_{\mathrm{RoI}}} \left( \lambda_p J_{\mathrm{nav}}(q_t,\tau) - \lambda_i I(\tau) \right),
    \label{eq:frontier_target}
\end{equation}
where $\mathcal{F}_{\mathrm{RoI}}$ is the set of frontier cells within the RoI, $I(\tau)$ is the expected information gain around the candidate cell $\tau$, and $\lambda_p, \lambda_i$ weight navigation cost and map expansion, respectively. Therefore, \eqref{eq:frontier_target} selects the best frontier cell to expand the map through low-cost terrain.

The selected target $\tau^\star$  is sent to the autonomous navigation system, implemented following the architecture in~\cite{macenski2020marathon2}. The planner computes a path over the cost representation derived from geometric traversability, and the path is tracked with a regulated pure-pursuit controller~\cite{macenski2023regulated}. The navigation controller outputs base velocity commands, which are sent to the locomotion controller and converted into joint-level torque commands applied to the robot. After each motion segment, the global elevation, traversability, navigation cost, and proprioceptive layers are updated, and new frontiers are extracted until no valid frontier remains inside the RoI or the desired coverage is reached.


\section{IMPLEMENTATION AND SYSTEM EVALUATION}
\subsection{Robotic System and Simulation Environment}

The proposed framework was implemented in ROS 2 and evaluated in NVIDIA Isaac Sim and OmniLRS~\cite{richardOmniLRS2024}. The simulator provides integration with ROS 2 and enables robot interaction with lunar-like scenarios. The tests were conducted under lunar gravity, with $g_{\mathrm{moon}} \approx 1.62~\mathrm{m/s^2}$, on a simulated lunar surface. The terrain presents elevation variations producing slopes, depressions, and uneven regions. The surface also includes scattered rocks with heights and diameters ranging from $0.1~\mathrm{m}$ to $0.6~\mathrm{m}$, as well as craters of up to $9.0~\mathrm{m}$ in diameter and $1.0~\mathrm{m}$ in depth.

\begin{figure}[htbp]
    \centering
    \includegraphics[width=0.99\linewidth]{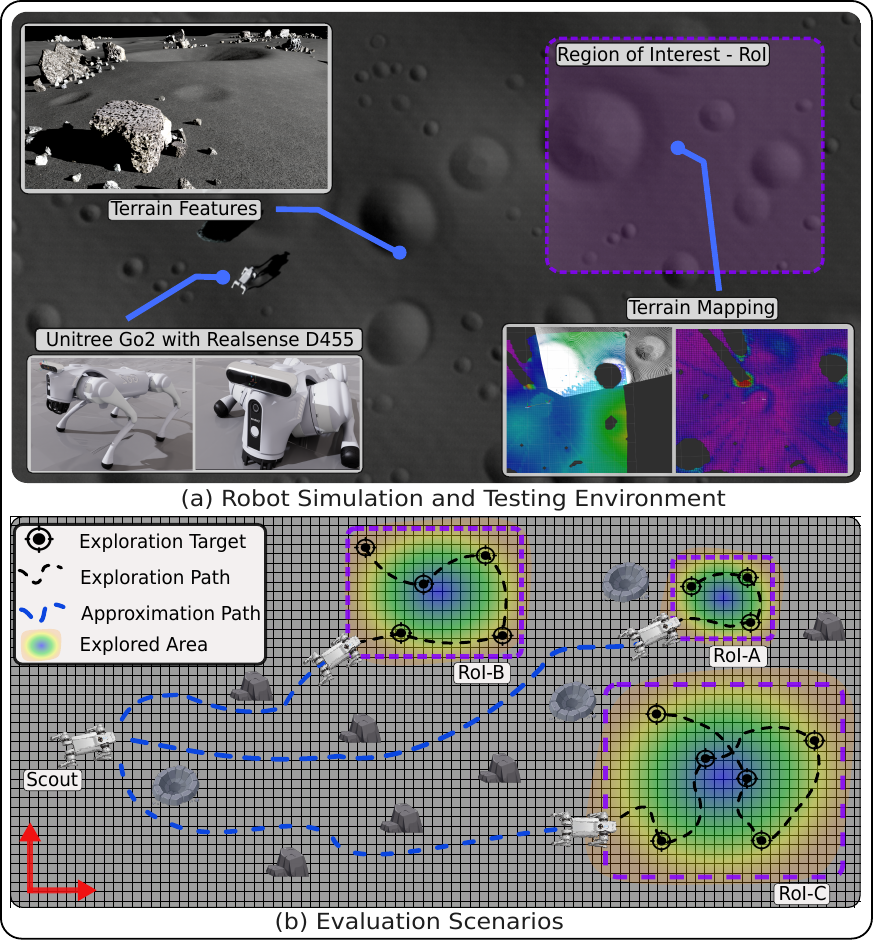}
    \caption{Implementation and evaluation of the system. 
        (a) Scout and simulation environment elements. 
        (b) Scenarios for evaluating the autonomous exploration.} 
    \label{fig:evaluation}
\end{figure}

The robotic platform is a simulated digital twin of the Unitree Go2 quadruped equipped with an Intel RealSense D455 camera. In the following, this robot is referred to as \textit{the scout}, since its role is to autonomously explore selected terrain regions and build terrain representations useful for navigation and assessment. The camera provides color images, depth images, and point clouds used as inputs for elevation mapping and geometric traversability estimation. In addition, the robot provides the proprioceptive signals required by the mapping framework, including joint states, inertial measurements from an IMU, and foot-contact information. Low-level locomotion control uses Quadruped-PyMPC~\cite{turrisi2024sampling}, an open-source Model Predictive Control framework. Figure~\ref{fig:evaluation}a shows the scout, the simulated lunar environment, and a representative RoI used during the evaluation.

\begin{figure*}[htbp]
    \centering
    \includegraphics[width=0.90\linewidth]{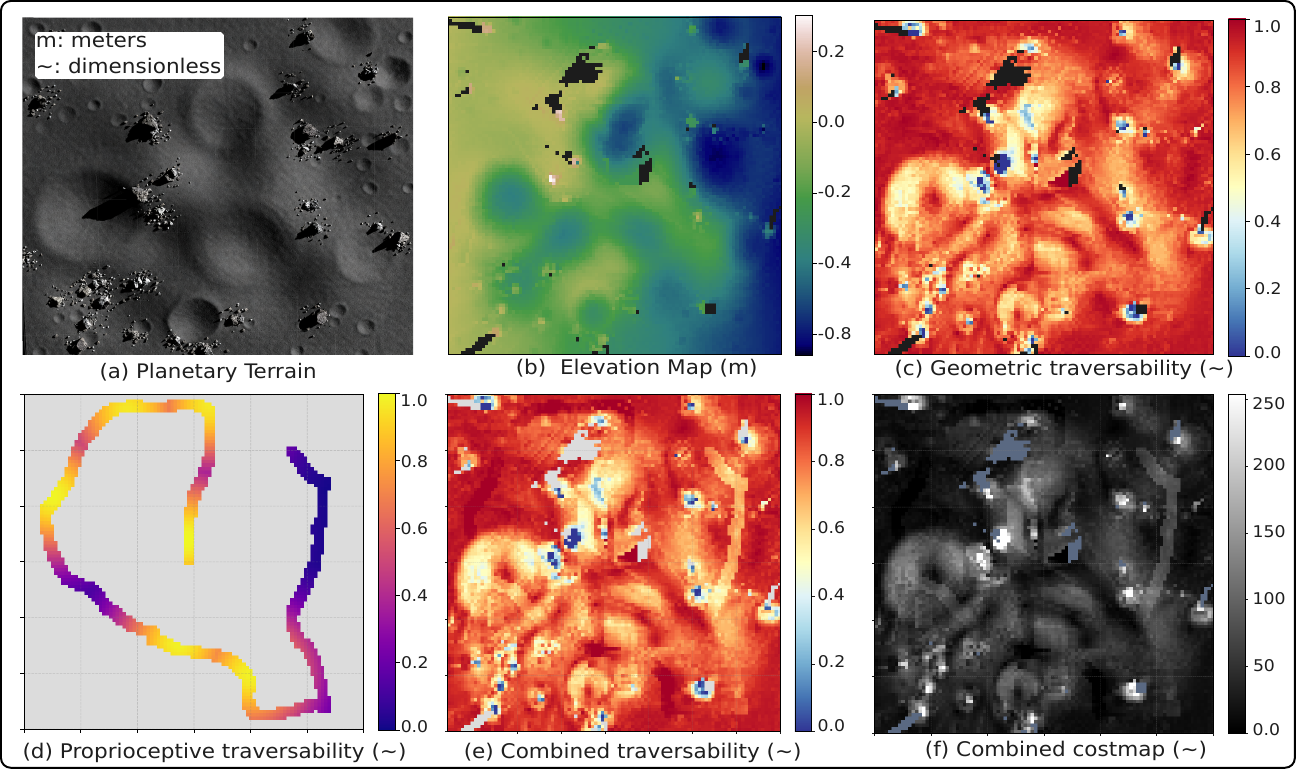}
    \caption{Global maps generated after exploration of a  RoI. 
        (a) Top-down view of the simulated lunar terrain. 
        (b) Elevation. 
        (c) Geometric traversability. 
        (d) Proprioceptive traversability. 
        (e) Combined traversability. 
        (f) Combined costmap.}
    \label{fig:roi_mapping}
\end{figure*}

\subsection{Autonomous Exploration Strategy}
The autonomous exploration process starts with the scout ready to walk on the simulated terrain. After receiving a request to explore a RoI defined in the global frame, the scout first approaches it through the autonomous navigation system. During this approximation phase, the robot follows a path from its initial position toward the RoI center, while avoiding obstacles detected along the way. This stage is illustrated in Fig.~\ref{fig:evaluation}b by the blue dashed path. This allows the scout to reach the target area even when it does not start inside the RoI.

Once in the RoI, the exploration module generates exploration targets to expand the explored area. These targets are sent sequentially to the navigation system: when the scout reaches one target, the next valid one is generated or selected according to the current exploration state. As new terrain is observed, the candidate targets are continuously updated, allowing the process to adapt to the partially built map. Exploration and mapping occur simultaneously; while the scout navigates between targets, the RGB-D camera point cloud updates the exteroceptive layers, while the proprioceptive measurements update the interaction-aware layers. All layers are integrated into the global multi-layer gridmap described in Section~II.

The process ends when no unexplored regions remain inside the RoI and all reachable targets have been visited. The output is a global representation of the explored area comprising elevation, geometric traversability, navigation cost, and proprioceptive traversability layers.
Figure~\ref{fig:evaluation}b shows a conceptual example for multiple RoIs, including the approximation path (blue dashed line), exploration targets (target symbols), executed exploration path (black dashed line), and accumulated explored areas (colored regions).

\subsection{Evaluation Protocol}

The evaluation was designed to assess the complete exploration and mapping pipeline under three RoIs of increasing size, as illustrated in Fig.~\ref{fig:evaluation}b. For each RoI size, tests were conducted under fixed initial conditions and identical navigation, locomotion, and mapping parameters, while only the RoI size was varied.

Each trial was executed until no valid reachable frontier remained inside the RoI, or until a maximum exploration budget of $1200~\mathrm{s}$ was reached. Explored coverage was computed as the percentage of cells inside the RoI that transitioned from unknown to observed in the global exteroceptive map. All layers were stored in a global gridmap with a spatial resolution of $0.1~\mathrm{m}$.

In parallel, the generated global maps were stored for subsequent analysis, including elevation, geometric traversability, navigation cost, and proprioceptive traversability layers. Additional proprioceptive indicators, such as CoT, were also logged to characterize the physical interaction and performance of the scout during exploration.



\section{RESULTS}
\subsection{Global RoI Mapping}

Figure~\ref{fig:roi_mapping} shows the maps generated during the autonomous exploration of a RoI. Figure~\ref{fig:roi_mapping}a shows the planetary terrain used in the trial. During the exploration, the scout traversed the selected region and the corresponding terrain layers were accumulated into the global map without relying on a prior terrain representation.

The RGB-D-based layers are shown in Fig.~\ref{fig:roi_mapping}b and Fig.~\ref{fig:roi_mapping}c. The elevation map captures the main height variations of the terrain, including depressions, rocks, and uneven regions. From this elevation layer, the geometric traversability map highlights areas with different terrain difficulty according to the local geometry. 

Figure~\ref{fig:roi_mapping}d shows the proprioceptive traversability layer. Unlike the exteroceptive layers, this map is populated only along the executed trajectory, where physical interaction with the terrain occurred. This spatial difference is expected, since exteroceptive information can be obtained from visible terrain around the robot, whereas proprioceptive information is only available over traversed cells. Finally, Fig.~\ref{fig:roi_mapping}e and Fig.~\ref{fig:roi_mapping}f show the combined traversability and costmap layers obtained for the same RoI. These maps summarize the terrain information accumulated during the exploration and provide the representation used in the subsequent navigation evaluation.

\subsection{Exploration Performance across RoI Scales}

\begin{figure}[H]
    \centering
    \includegraphics[width=0.90\linewidth]{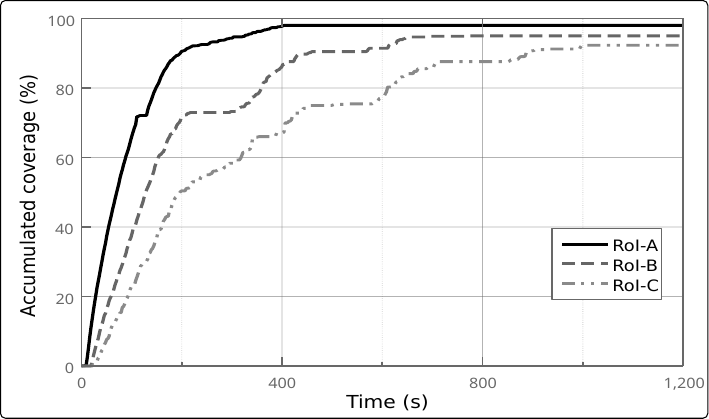}
    \caption{Coverage during autonomous exploration.}
    \label{fig:coverage}
\end{figure}

\begin{figure}[H]
    \centering
    \includegraphics[width=0.90\linewidth]{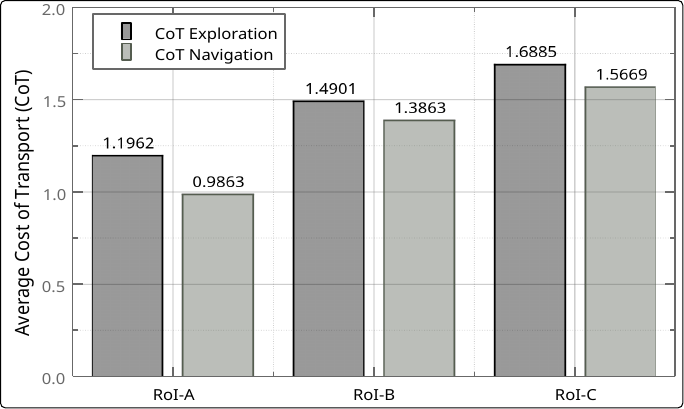}
    \caption{Average CoT during exploration and subsequent navigation.}
    \label{fig:CoT}
\end{figure}

Figures~\ref{fig:coverage} and \ref{fig:CoT} summarize the exploration performance for RoIs of increasing size. The same exploration, navigation, and mapping parameters were used in all cases, while only the RoI area was varied. As shown in Fig.~\ref{fig:coverage}, the explored coverage increases monotonically in the three RoIs. The smallest region reaches high coverage in a shorter time, whereas larger RoIs require longer exploration and show a more gradual convergence. This behavior is consistent with the need to generate additional targets and progressively extend the global map over wider areas.

The final exploration metrics are reported in Table~\ref{tab:exploration_metrics}. As the RoI size increases, the scout requires longer execution time, longer traveled distance, and a larger number of exploration targets before the process saturates. The final coverage remains above 94\% in all cases, although it decreases slightly for larger RoIs. This reduction is consistent with the larger area to be covered and with regions that may remain partially observed or difficult to reach during autonomous exploration.

\begin{table}[ht]
    \centering
    \caption{Exploration metrics for different RoI scales}
    \label{tab:exploration_metrics}
    \begin{tabular}{llccccc}
    \hline
    RoI & Size & Area & Coverage & Time & Distance & Targets \\
        &      & (m$^2$) & (\%) & (s) & (m) & -- \\
    \hline
    RoI-A & Small  & 400  & 97.72 & 430 & 210 & 6  \\
    RoI-B & Medium & 900  & 95.04 & 620 & 305 & 10  \\
    RoI-C & Large  & 1600 & 94.68 & 820 & 410 & 14 \\
    \hline
    \end{tabular}
\end{table}

Figure~\ref{fig:CoT} compares the average CoT measured during exploration with a subsequent navigation stage over the same target sequence. In this second stage, the scout reuses the terrain information accumulated during exploration, including the global traversability and cost representations. The average CoT is lower during the navigation stage for the three RoIs. This result suggests that the accumulated map can provide useful information for revisiting previously explored targets with lower locomotion cost, although additional trials would be required to isolate the contribution of each map layer and navigation decision.

\section{DISCUSSION AND LIMITATIONS}
The results show that the framework can autonomously explore bounded RoIs while building a multi-layer map that relates geometric terrain structure with robot-terrain interaction. The combined traversability layer should be interpreted as a lightweight experience-aware correction of the geometric map, rather than an optimal sensor-fusion method. The lower CoT observed during subsequent navigation suggests that the accumulated map can provide useful mobility information when revisiting previously explored targets.

The resulting cost should not be interpreted as morphology-invariant: different rovers may assess the same terrain cell differently depending on their size, mass, contact mechanics, and locomotion strategy. Nevertheless, the scout map provides localized terrain evidence for later agents by identifying areas associated with high geometric cost, slippage, instability, or energetic demand. These layers can therefore support risk-, safety-, and efficiency-aware route selection when interpreted together with the mobility constraints of each rover, rather than used as a directly transferable traversability and cost.

However, several limitations remain. Proprioceptive coverage is sparse and path-dependent, with potential reuse during return trips or repeated site visits. Our evaluation uses simulator ground-truth poses and does not assess localization uncertainty, which can misalign stored measurements with revisited terrain. Reuse requires consistent localization in the map frame; place recognition or loop closure can help but is not inherently required. The evaluation is also limited to simulation and a small number of RoIs, without statistical analysis, a full ablation separating the contribution of each layer, or a terrestrial-gravity baseline, to isolate the impact of reduced gravity on system performance. Finally, the simulated environment does not fully capture real lunar or analog-terrain effects such as deformable regolith, dust, extreme illumination, depth-sensing degradation, or hardware and locomotion uncertainties.

\section{CONCLUSIONS AND FUTURE WORK}
This paper presented a terrain-aware autonomous exploration framework for quadruped scouts in lunar-like environments. The proposed system integrates RGB-D-based elevation mapping, geometric traversability estimation, proprioceptive terrain assessment, frontier-based target generation, and autonomous navigation into a unified pipeline. Simulation results showed that the scout can explore RoIs of increasing size while incrementally building global multi-layer terrain maps. Subsequent navigation over the same target sequence using the combined traversability and cost maps exhibited lower average CoT than initial exploration. However, differences in map availability and exploration behavior prevent isolating the contribution of proprioceptive information beyond geometry. Future work will focus on validating the framework with hardware experiments and planetary analog terrains. Additionally, we plan to perform simulated ablation studies across diverse RoI scales and locations to isolate the contribution of each terrain layer, while improving the exteroceptive-proprioceptive combination through uncertainty-aware fusion or learned propagation of proprioceptive information.

\bibliographystyle{IEEEtran}
\bibliography{main}

\end{document}